\documentclass[pdflatex,sn-mathphys-num]{sn-jnl}% Math and Physical Sciences Numbered Reference Style
\usepackage{graphicx}%
\usepackage{multirow}%
\usepackage{amsmath,amssymb,amsfonts}%
\usepackage{amsthm}%
\usepackage{mathrsfs}%
\usepackage[title]{appendix}%
\usepackage{xcolor}%
\usepackage{textcomp}%
\usepackage{manyfoot}%
\usepackage{booktabs}%
\usepackage{algorithm}%
\usepackage{algorithmicx}%
\usepackage{algpseudocode}%
\usepackage{listings}%
\usepackage{hyperref}
\usepackage{array}

\theoremstyle{thmstyleone}%
\theoremstyle{thmstyletwo}%

\theoremstyle{thmstylethree}%

\begin{document}

\title[Bridging Minds and Machines]{\,\,\,\,\,\,\,\,\,\,\,\,\,\,\,\,\,\,\,Bridging Minds and Machines: \\Toward an Integration of AI and Cognitive Science}
%\title[The Conjunctions of Artificial Intelligence and Cognitive Science]{The Conjunctions of Artificial Intelligence and Cognitive Science: A Review}

%%=============================================================%%
%% GivenName	-> \fnm{Joergen W.}
%% Particle	-> \spfx{van der} -> surname prefix
%% FamilyName	-> \sur{Ploeg}
%% Suffix	-> \sfx{IV}
%% \author*[1,2]{\fnm{Joergen W.} \spfx{van der} \sur{Ploeg} 
%%  \sfx{IV}}\email{iauthor@gmail.com}
%%=============================================================%%

\author[1]{\fnm{Rui} \sur{Mao}}\email{rui.mao@ntu.edu.sg}

\author[2]{\fnm{Qian} \sur{Liu}}\email{liu.qian@auckland.ac.nz}

\author[3]{\fnm{Xiao} \sur{Li}}\email{xiao.li@abdn.ac.uk}

\author*[1]{\fnm{Erik} \sur{Cambria}}\email{cambria@ntu.edu.sg}

\author[4]{\fnm{Amir} \sur{Hussain}}\email{a.hussain@napier.ac.uk}

\affil[1]{\orgname{Nanyang Technological University}, \orgaddress{\country{Singapore}}}

\affil[2]{\orgname{University of Auckland}, \orgaddress{\country{New Zealand}}}

\affil[3]{\orgname{University of Aberdeen}, \orgaddress{\country{United Kingdom}}}

\affil[4]{\orgname{Edinburgh Napier University}, \orgaddress{\country{United Kingdom}}}

%%==================================%%
%% Sample for unstructured abstract %%
%%==================================%%

%\abstract{Cognitive Science, as a fundamental discipline, has deeply influenced the development of broad fields, such as Artificial Intelligence (AI), Philosophy, Psychology, Neuroscience, Linguistics, and Culture. Recent advancements in AI were largely inspired by cognitive theories; meanwhile, AI could also be used in cognitive analysis as a powerful tool. These mutual benefits motivated us to review the conjunctions of AI and other research fields in Cognitive Science. By outlining significant research outcomes from both perspectives, we find that the advancements of AI have largely focused on practical task processing, whereas the cognitive grounding has remained conceptually fragmented. We believe that the future of AI in Cognitive Science depends not merely on performance gains but on building systems that extend our deepest understanding of the human mind. Thus, a few directions deserve more attention, e.g., aligning AI behavior with cognitive frameworks; embedding AI in embodiment and culture; developing personalized cognitive representations; and re-conceptualizing AI ethics through cognitive co-evaluation.}
\abstract{Cognitive Science has profoundly shaped disciplines such as Artificial Intelligence (AI), Philosophy, Psychology, Neuroscience, Linguistics, and Culture. Many breakthroughs in AI trace their roots to cognitive theories, while AI itself has become an indispensable tool for advancing cognitive research. This reciprocal relationship motivates a comprehensive review of the intersections between AI and Cognitive Science. By synthesizing key contributions from both perspectives, we observe that AI progress has largely emphasized practical task performance, whereas its cognitive foundations remain conceptually fragmented. We argue that the future of AI within Cognitive Science lies not only in improving performance but also in constructing systems that deepen our understanding of the human mind. Promising directions include aligning AI behaviors with cognitive frameworks, situating AI in embodiment and culture, developing personalized cognitive models, and rethinking AI ethics through cognitive co-evaluation.}


%%\pacs[JEL Classification]{D8, H51}

%%\pacs[MSC Classification]{35A01, 65L10, 65L12, 65L20, 65L70}

\maketitle

\section{Introduction}\label{sect: Introduction}

\begin{figure}[!t]
    \centering
    \includegraphics[width=1\linewidth]{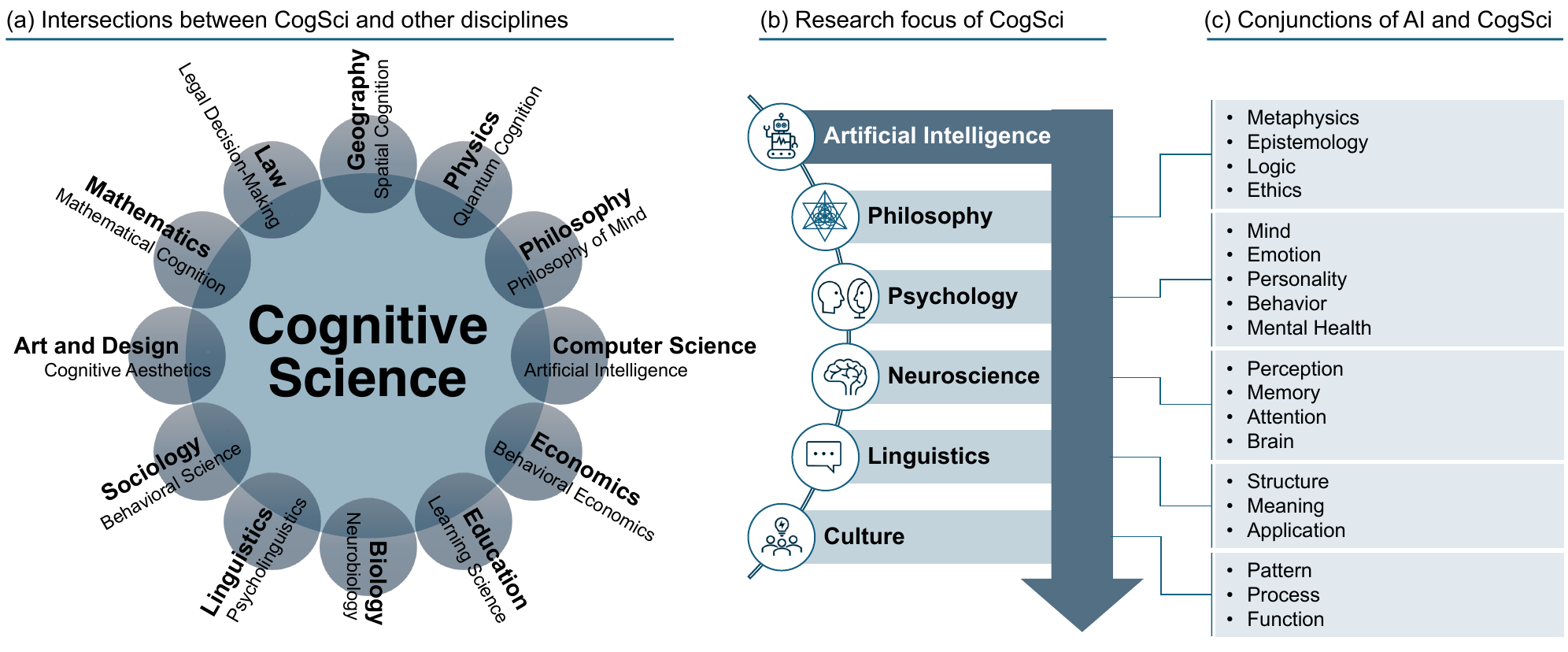}
    \caption{The relationship between AI and Cognitive Science (CogSci).}
    \label{fig:relationship}
\end{figure}

Cognitive Science is an important field of research dedicated to understanding the human mind and its cognitive processes~\citep{willingham2005ask}. As seen in Fig.~\ref{fig:relationship}.a, it demonstrates extensive interdisciplinary connections, leading to the emergence of new research areas that integrate cognitive principles with various other disciplines. For example, the intersection of Cognitive Science and Computer Science has given rise to Artificial Intelligence (AI) that enables machines to mimic how people think and behave~\citep{lecun2015deep}. Similarly, Behavioral Economics has emerged from the convergence of Cognitive Science and Economics by employing cognitive frameworks to analyze economic patterns~\citep{thaler2016behavioral}. The broad connections show the significance of Cognitive Science in advancing different scientific disciplines.

We adopt the classification scheme outlined in \textit{The MIT encyclopedia of the Cognitive Sciences}~\citep{wilson2001encyclopedia}, which surveys representative contributions from philosophy, psychology, neuroscience, computational intelligence, linguistics, and cultural studies. This framework closely aligns with the \textit{Cognitive Science Hexagon}~\citep{milliere2024philosophy}, with the primary difference being the substitution of culture with anthropology. These research fields leverage different methods to understand humans' thoughts, feelings, and behaviors. For example, philosophy studies fundamental questions about consciousness and the self; psychology provides empirical methods to analyze observable behavior and cognitive processes; neuroscience looks under the hood, exploring the neural basis of thoughts and behaviors; computational intelligence simulates cognitive processes and explores human-like intelligence in artificial systems. Linguistics examines the relationship between language and cognition; cultural studies reflect how cognition is influenced by social factors. Together, these fields offer a layered understanding of what it means to be an intelligent being. 

AI\footnote{We consider terms, AI and Computing Intelligence, as synonyms, given the shared focus on creating systems that emulate human cognitive capabilities and the broad spectrum of computational methods encompassed by AI.} is a Computer Science research field focused on creating algorithms that allow machines to perceive the environment, learn, and make decisions to achieve goals~\citep{russell2016artificial}. The relationship between AI and Cognitive Science is interwoven (see Fig.~\ref{fig:relationship}.b). The technical development of AI has been frequently inspired by cognitive theories, such as attention mechanisms~\citep{vaswani2017attention} and long short-term memory (LSTM)~\citep{hochreiter1997long}. On the other hand, the advancement of AI can also provide useful tools for cognitive analysis and pattern recognition. With the development of large language models (LLMs) such as ChatGPT~\citep{openai2023chatgpt}, Generative AI (GenAI) has been widely used in different scientific research fields~\citep{mao2024gpteval}. Thus, we are motivated to review the conjunction of AI and Cognitive Science in this work. The AI research included in this work serves two primary purposes: either demonstrating application values within Cognitive Science (AI for cognitive analysis), or driving advancements in AI research methodologies through the lens of cognitive tasks and theories (cognition-inspired AI).

The specific AI research areas that are covered by this work are detailed in Fig.~\ref{fig:relationship}.c. In contrast to recent surveys grounding AI in particular cognitive theories or domains~\citep{langley2022theory, achler2024ai, van2024reclaiming}, we aim to provide a review of state-of-the-art AI technologies that are relevant to the critical concerns across the diverse fields of Cognitive Science. Additionally, this work also explores the research gap between current AI development and the theoretical foundations of Cognitive Science. We believe that insights from this comparative analysis can stimulate more expansive and profound AI research, enabling AI to not only achieve human-level accuracy in task processing but also to develop human-like intelligence in perceiving the world.

\section{Philosophy}\label{sec: Philosophy}

Philosophy revolves around fundamental questions about existence, knowledge, values, reason, mind, and language~\cite{deleuze1994philosophy}. Distinguished by its unique methodologies and expansive subject matter, philosophy tackles questions that touch on every aspect of our lives. In philosophy research, the representative topics are commonly taken to be metaphysics, epistemology, logic, and ethics. 
%Additionally, the philosophy of mind and language constitutes significant branches of investigation. Given the particular relevance of mind and language to the following sections, these topics are not discussed here.

\begin{figure}[!t]
    \centering
    \includegraphics[width=1\linewidth]{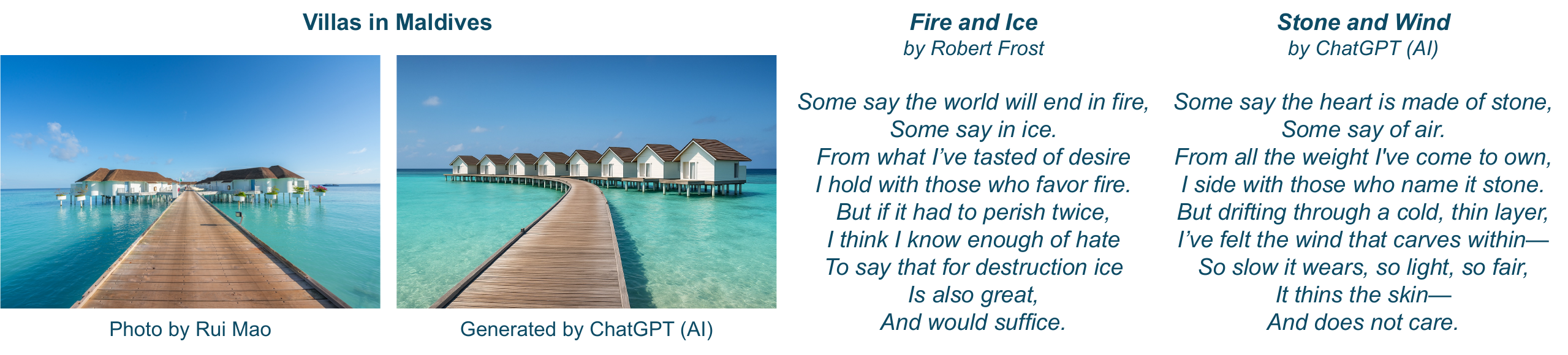}
    \caption{A comparison between AI-generated content and their human-crafted counterparts.}
    \label{fig:gen_ai}
\end{figure}

\textbf{Metaphysics} is the study of reality, existence, and the universe~\cite{carroll2010introduction}. It tries to understand existence, objects and their properties, wholes and their parts, space and time, events, and causation~\cite{VanInwagen2023-VANMQA-3}. Technical progress in Virtual Reality (VR), and Augmented Reality (AR)~\citep{rauschnabel2022xr} blurs the boundary between the physical and the digital, raises philosophical questions about what makes an experience ``real'' versus ``simulated''~\citep{bojic2022metaverse}. Similarly, questions regarding AI consciousness~\citep{dehaene2021consciousness}, autonomy~\citep{formosa2021robot}, and the ontological significance of content, generated by AI~\cite{o2025artificial} (see examples in Fig.~\ref{fig:gen_ai} and statements below\footnote{The left photo was personally taken and does not violate any third-party IP rights. The AI-generated content is from ChatGPT (\url{https://chatgpt.com/}), following OpenAI's usage policies (\url{https://openai.com/zh-Hant/policies/usage-policies/}). Prompt for the image generation: Generate a high-resolution, horizontal (6:9 aspect ratio), realistic Maldives-style photo, featuring a clear blue sky, turquoise sea, a cluster of white overwater villas with brown roofs, and a wooden boardwalk extending over the water leading toward the villas. Prompt for the poem generation: Create a poem inspired by the style and themes of Robert Frost's ``Fire and Ice''~\citep{frost1920fire}.}) lead us to rethink long-standing metaphysical concepts. However, AI research typically takes ontological categories for granted without looking at how they are based. AI can hardly understand the existential meaning, intentionality, and reality status of the very entities it seeks to generate.

\textbf{Epistemology} studies the nature, scope, and acquisition of knowledge, addressing questions about what we know and how we come to know it~\cite{CrumleyIi2009-CRUAIT}. It defines knowledge, exploring knowledge origins, boundaries, and significance. Rooted in this, knowledge-based systems (KBS) use explicit knowledge bases and inference engines~\citep{akerkar2009knowledge}, e.g., Wikipedia\footnote{\url{https://www.wikipedia.org/}}, often supported by ontologies (e.g., WordNet~\citep{miller1995wordnet}), to formalize and share knowledge with symbolism. However, recent methods, based on neural networks and deep learning, have shifted knowledge representation, particularly for semantic knowledge, towards embeddings (e.g., word2vec~\citep{mikolov2013word2vec}) and pre-trained language models (e.g., BERT~\citep{devlin2019bert}). This change promotes data-driven, distributed representations over symbolic rules, enabling AI to capture complex semantic relationships and generalize from big data. However, current AI research largely bypasses normative concerns, privileging statistical correlation over epistemic warrant, and thereby risks conflating information processing with genuine knowledge acquisition or understanding.

\textbf{Logic} is the study of valid reasoning, distinguishing sound from unsound arguments~\cite{Quine1986-QUIPOL-7}, and encompasses both formal and informal branches. In AI research, Prolog-based systems~\citep{korner2022fifty} use formal logic, e.g., first-order logic (FOL)~\citep{han2024folio}, to automatically prove or disprove logical statements. Informal logic directs AI systems to process uncertainty~\citep{talpur2023deep}, natural language~\citep{manning1999foundations}, or human-like reasoning~\citep{hagendorff2023human} through cognitive and probabilistic approaches, including Natural Language Understanding (NLU)~\citep{cambria2024nlu}, Probabilistic Logic (e.g., Chain-of-Thought (CoT)~\citep{wei2022chain}, Case-Based Reasoning (CBR)~\citep{kolodner2014case}, and Commonsense Reasoning (CSR)~\citep{davis2015commonsense}. However, AI research often operationalizes logic as a functional tool~\citep{shen2025flow}, raising concerns about the normativity of inference, and the awareness that distinguishes merely plausible outputs from genuinely sound reasoning.

\textbf{Ethics} explores questions related to moral values and principles, e.g., what is right and wrong, what are the moral standards, and how to live a good life~\citep{Hagendorff2020-HAGTEO-9}. In AI, these ethical concerns center on its fairness, accountability, and transparency~\citep{shin2019role}. Biases are frequently analyzed from the perspectives of algorithms~\citep{mao2023biases} and language~\citep{ge2025toxicbias}. To align AI systems with human values, reinforcement learning from human feedback (RLHF) was employed for LLM fine-tuning~\citep{ouyang2022training}. Explainable AI (XAI) becomes a new trend in building trustworthy and transparent AI~\citep{camxai}, particularly in high-stakes tasks such as healthcare~\citep{han2022hierarchical,he2025survey} and finance~\citep{yeo2025survey}. However, current AI ethics frequently focus on immediate concerns, overlooking long-term implications for human society, such as how to preserve a meaningful life in a world increasingly influenced by AI systems; does the purpose of creating AI lies in the displacement of human labor, or in the augmentation of human endeavor.

\section{Psychology}\label{sec: Psychology}

%Psychology is the scientific study of the mind and behavior. Although mutually informative within the broader field of Cognitive Science, psychology and neuroscience differ in their primary orientations and methodologies. Psychology emphasizes mental processes, affective experiences, personality traits, behavioral patterns, and mental health at the level of individuals and groups. It uses methods, e.g., behavioral experiments, surveys, clinical interviews, and observational studies to understand how people think, feel, and act. Neuroscience, in contrast, examines the biological mechanisms underlying these phenomena, studying the brain and nervous system via techniques such as brain imaging, electrophysiology, and molecular biology. Although both disciplines may explore domains such as perception, memory, attention, and brain function, they approach these topics from distinct yet complementary vantage points. To clarify the neuroscientific contributions to Cognitive Science and distinguish them from psychological perspectives, we introduce these domains from a neuroscience standpoint in the next section. In this section, we focus on mind, emotion, personality, behavior, and mental health, which are representative topics in psychology.

Psychology is the scientific study of the mind and behavior, including mental processes, affective experiences, personality traits, behavioral patterns, and mental health at the level of individuals and groups. Although psychological research frequently employs neuro-imaging methods, this section focuses on representative topics from behavioral and cognitive perspectives to distinguish it from the next section.

\textbf{Mind} refers to the sum of an individual's conscious and unconscious mental processes and states~\citep{bargh2008unconscious}. Theory of Mind (ToM) believes that individuals have the cognitive capacity to attribute mental states, e.g., beliefs, desires, intentions, and emotions, to themselves and others~\citep{apperly2009humans}. Children gain ToM capacities around the age of four~\citep{wellman2001meta}. In a large-scale comparison involving 1,907 human participants, GPT-4 matched or exceeded humans on ToM assessments of false beliefs, indirect requests, and misdirection, though it struggled with detecting faux pas~\citep{strachan2024testing}. Another research finds that LLMs' false-belief test performance is similar to 6-year-old children~\citep{kosinski2024evaluating}. However, AI-based ToM research remains limited in scope and explanatory depth. Unlike humans, GenAI lacks subjective awareness and intentional understanding, which are the core features of the psychological concept of mind. Their performance is evaluated via observable behavior alone, which risks conflating surface-level mimicry with genuine mental reasoning. 

\textbf{Emotion} psychology studies how emotions are generated~\citep{scherer2005emotions}, experienced~\citep{barrett2007experience}, regulated~\citep{mega2014makes}, and expressed~\citep{ekman2014expression} across diverse contexts. Psychologists explored basic emotions~\citep{ekman2000basic} and uncovered the cognitive processes to explain humans' emotional reactions~\citep{moors2013appraisal,hourglass}. Affective computing, on the other hand, primarily focuses on detecting~\citep{fan2024fusing} and simulating emotional expressions~\citep{hu2024human} from surface-level signals, e.g., facial movements~\citep{yang2024robust}, vocal tone~\citep{chen2024vesper}, or textual sentiment~\citep{cambria2024senticnet}. While these systems can classify emotions with higher accuracy and finer granularity, their analytical scope remains in pattern recognition without the mechanism of modeling the mental processes of emotion. The limited convergence between psychological and computational approaches highlights the need for more integrative frameworks that bridge expressive patterns with underlying emotional mechanisms.

\textbf{Personality} psychology studies individual differences in traits and how they influence behavior. The Big Five model (OCEAN)~\cite{roccas2002big} and Myers-Briggs Type Indicator (MBTI)~\cite{myers1962myers} are widely used frameworks for categorizing personality traits. Empirical studies evaluate the personality traits in different contexts, such as job performance~\citep{hurtz2000personality}, education~\citep{harrington2010mbti}, and financial risk-taking~\citep{filbeck2005risk}, although the psychometric limitations of these models were frequently criticized~\citep{furnham1996big,gosling2003very}. AI-based personality trait recognition (PTR)~\citep{li2022multitask} and persona extraction from text~\citep{zhu2023pead} are necessary for human-computer interaction, empowering personalized systems across diverse contexts, e.g., precision medicine~\citep{wu2025ehr}, affective computing~\citep{zhu2024neurosymbolic,xie2025pgif}, adaptive dialogue systems~\citep{zhu2024hippl}, and recommender systems~\citep{zhang2024generative}. However, similar to the critique from psychologists, a critical challenge lies in the limited ecological and cultural validity of AI systems. Besides, in order to model human diversity, there is a need for new technology and knowledge to understand personality.

\begin{figure}[!t]
    \centering
    \includegraphics[width=1\linewidth]{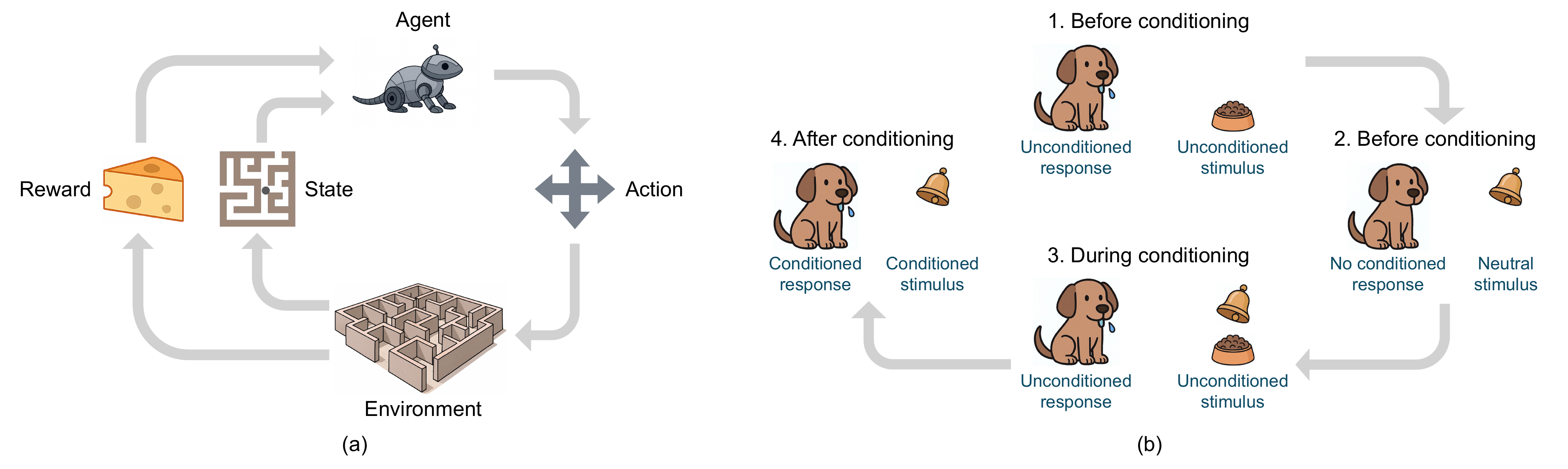}
    \caption{(a) Reinforcement Learning~\citep{sutton1998reinforcement}: an agent interacts with the environment, performs actions, and receives rewards or penalties. (b) Pavlov's Dogs experiment~\citep{pavlov2010conditioned}: dogs were trained to link a neutral stimulus (e.g., a bell) with food, causing them to salivate at the bell alone. Both reflect the concept of learning through experience in behavioral science.}
    \label{fig:behavior}
\end{figure}

\textbf{Behavioral studies} in psychology explore the relationship between observable behavior and external factors, including learning through experience~\citep{pavlov2010conditioned}, habit development by repetition~\citep{gardner2022does}, decision-making influenced by past experiences~\citep{klein2017sources}, and expectations~\citep{granovetter1973strength}, as well as behavioral interventions by external incentives~\citep{michie2011behaviour}. In AI, robotics research~\citep{brunke2022safe} has frequently integrated concepts from behavioral science, particularly through Reinforcement Learning (RL, see Fig.~\ref{fig:behavior})~\citep{sutton1998reinforcement}. Social Network Analysis (SNA)~\citep{freeman2004development} and sentiment analysis-based financial prediction~\citep{du2024financialsentiment} quantify social impacts on decision making. AI-driven applications have been also used for smoking cessation~\citep{bendotti2023conversational} and fitness~\citep{farrokhi2021application}. However, AI models often assume optimized behavior, overlooking the individual irrationality and emotional complexity that define real human actions~\citep{simon1955behavioral}. These are not flaws but essential characteristics of human behavior that must be acknowledged rather than abstracted away.

\textbf{Mental health} is frequently studied in behavioral science, emotion theory, and neuro-imaging. However, what distinguishes the field are its specialized domains such as prevention~\citep{haggerty1994reducing}, diagnostics~\citep{kawa2012brief}, psychotherapy~\citep{olfson2025psychotherapy}, trauma and resilience~\citep{bonanno2011resilience}, and mental health across the lifespan~\citep{oerlemans2020association}, involving highly personalized, and culture-contextualized human experiences. Using AI to detect mental health detection~\citep{zhang2022natural,nielsen2020machine} is vivid, where XAI contributes by explaining the underlying cognitive representations and how AI models reach their decisions~\citep{han2022hierarchical}. Researchers have also begun to examine the psychological safety of GenAI, especially concerning their potential to produce misleading effects~\citep{li2025gaslighter}. While these directions are encouraging, mental health is more than identifying disorders. AI should support deeper therapeutic engagement, enhance self-awareness, and promote psychological growth.

\section{Neuroscience}\label{sec: Neuroscience}

%Neuroscience aims to understand how the brain and its networks give rise to complex cognitive functions, including perception, attention, memory, emotion, and behavior. By integrating approaches at the molecular, cellular, systems, and cognitive levels, it aims to reveal the neural mechanisms that underlie mental processes and mediate responses to both internal states and external environments. To achieve this, neuroscience employs a wide range of research methods, including neuro-imaging techniques (e.g., fMRI, PET, EEG, MEG) for visualizing brain activity in humans, electrophysiology for recording neuronal signals, and molecular and genetic tools (e.g., optogenetics, CRISPR) for targeted manipulation of neural circuits. Psychological and behavioral experiments further connect neural activity to observable cognition and actions, while studies in both animal models and humans offer complementary insights into foundational mechanisms and clinical applications. The following content in this section explores how neuroscience and AI intersect in understanding core cognitive domains, such as perception, memory, attention, and brain function.

Neuroscience studies how the brain and its networks enable complex cognitive functions, such as perception, attention, memory, emotions, and behavior. By integrating research methods at the molecular, cellular, systemic, and cognitive levels, it aims to discover the neural mechanisms that constitute psychological processes and mediate responses to internal states and external environments.

\begin{figure}[!t]
    \centering
    \includegraphics[width=1\linewidth]{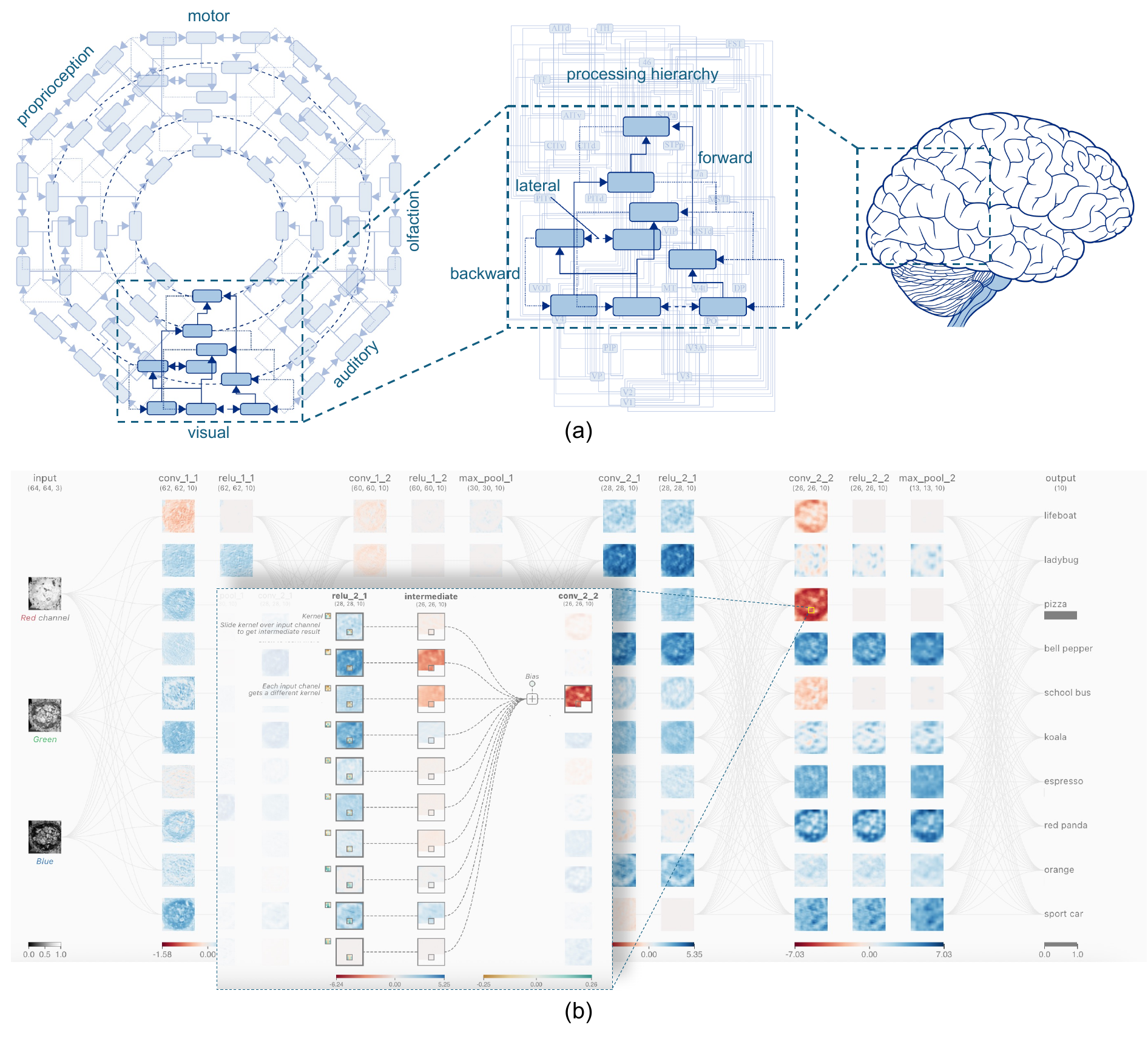}
    \caption{(a) The brain's perception hierarchy~\citep{friston2005theory}, based on sensory-fugal processing along a core synaptic hierarchy that spans primary sensory areas through unimodal, heteromodal, paralimbic, and limbic regions of the cerebral cortex~\citep{mesulam1998sensation}. Reproduced. (b) A typical convolutional neural network that extracts hierarchical features from the input and outputs a prediction — pizza. Visualized with CNN Explainer~\citep{wangCNNExplainerLearning2020}).}
    \label{fig:perception}
\end{figure}

\textbf{Perception} refers to the ability that brain interprets sensory input and builds up connections to external environments. To perceived and process signals from vision~\citep{britten1996relationship}, hearing~\citep{belin2000and}, touch~\citep{gordon2013brain}, taste~\citep{wang2004taste}, and smell~\citep{shepherd2006smell}, complicated neural computation across sensory and cortical regions~\citep{harris2013cortical} is required. The brain develops a perception hierarchy to dynamically reconcile discrepancies between top-down predictions and bottom-up error signals~\citep{friston2005theory}, enabling internal models to be updated from real-time input. The layered structure is also manifested in deep neural networks (DNNs) in AI (see comparison in Fig.~\ref{fig:perception}). For example, researchers developed convolutional neural networks (CNNs)~\citep{krizhevsky2012imagenet} with multi-layers. The lower layers extract simple features; the higher layers capture abstract patterns. In contrast to human's capacity that can learn knowledge from a few samples, current DNNs require a lot of samples for training. This is a significant limitation of current AI perception methods.

\textbf{Memory} is more than just keeping track of information. It also includes encoding, consolidation, retrieval, and forgetting mechanisms. Sensory input is first retained in modality-specific buffers (e.g., iconic and echoic memory~\citep{coltheart1980iconic,naatanen1989event}) for milliseconds to seconds~\citep{di1977temporal,jaaskelainen1999temporal}. Attended information then goes in working memory~\citep{baddeley2020working}. Long-term storage will change neural connectivity~\citep{escobar2007long}. Over time, memories transfer to cortical networks through systems consolidation~\citep{squire2015memory}, while synaptic consolidation stabilizes memory traces at the cellular level~\citep{bailey2015structural}. Without reactivation, memory traces will be weakened because of synaptic downscaling~\citep{dudai2004neurobiology}. In contrast, AI stores knowledge in network weights. Models like LSTM~\citep{hochreiter1997long} and GRU~\citep{cho2014learning} simulate working memory via gating mechanisms. Since static knowledge can be retained on storage devices, retrieval-augmented generation systems combine LLMs with real-time retrieval to handle dynamic content~\citep{lewis2020retrieval}. However, current AI memory lacks capabilities such as spontaneous rehearsal, selective forgetting, and adaptive generalization across contexts.

\textbf{Attention} allows the brain to prioritize important sensory input or internal thoughts and filter out the insignificant ones. This resource allocating process is coordinated by distributed neural systems, including the frontoparietal network for voluntary (endogenous) attention and the ventral attention network for involuntary (exogenous) reorienting~\citep{corbetta2008reorienting}. Given the significance of attention in task processing, it was also introduced in machine translation~\citep{bahdanau2015neural} to address the challenge of encoding lengthy source language. Later, the Transformer architecture introduced multi-head self-attention~\citep{vaswani2017attention}, enabling models to attend to information from different vector spaces. However, the AI attention mechanisms are formed, typically based on the learning of statistical associations between targets and contexts. Uncommon context and target associations can result in hallucinated predictions.

\textbf{Brain} research is typically grounded in the domains of emotion, decision-making, motor control, and consciousness, using neuro-imaging techniques to understand brain activities in those contexts. Researchers found that emotional processing involves regions such as the amygdala, insula, and prefrontal cortex~\citep{menon2010saliency}, while decision-making engages the orbitofrontal cortex, basal ganglia, and dopaminergic systems to evaluate options and outcomes~\citep{schultz1997neural,padoa2006neurons}. Motor control relies on the motor cortex, cerebellum, and spinal circuits to coordinate movement~\citep{li2015motor}. Mirroring the functional specialization in the human brain, agentic AI is designed for tasks such as perception, planning, and decision-making~\citep{hongmetagpt}, achieving great independent task-processing capacities in dynamic environments. AI is also used for brain signal denoising~\citep{mishro2021survey}, and interpretation~\citep{wang2025explainable}. However, our knowledge about the reasoning mechanism of neural networks is still limited. Giving AI excessive autonomy and resource mobilization authority likely leads to serious consequences.

\section{Linguistics}\label{sec: Linguistics}

%Linguistic research aims to understand how language is structured (phonology, morphology, syntax), how meaning is constructed and interpreted (semantics, pragmatics), and how language is acquired, produced, comprehended, and represented in the mind and brain (applied linguistics)~\citep{jackendoff2002language}. Given the shared aim of modeling language understanding and generation, linguistic theory and computational linguistics are closely aligned in their core research tasks. Both disciplines study the mental mechanisms underlying language, though they differ in their methodological approaches: linguistics primarily employs formal grammars and representational analyses~\citep{hawkins2004efficiency}, while computational linguistics leverages algorithmic architectures and data-driven models~\citep{mitkov2022oxford}. Within AI, a closely related field is natural language processing (NLP). Although NLP and computational linguistics often overlap~\citep{hirschberg2015advances}, especially in their use of computational methods, we draw a distinction between the two in this work: NLP is primarily concerned with building systems to perform practical, language-related tasks, whereas computational linguistics focuses more directly on modeling the structure and function of language itself. The following AI review is from the perspective of computational linguistics, rather than NLP. Thus, the reviewed works exclude conventional NLP tasks, e.g., dialogue systems, text summarization, and other practical language tasks.

Linguistic research aims to understand how language is structured (phonology, morphology, syntax); how meaning is constructed (semantics, pragmatics); and how language is acquired, and represented in mind (applied linguistics)~\citep{jackendoff2002language}. Given the shared aim of modeling language understanding and acquisition, linguistics and computational linguistics are well aligned in their core research tasks. 

\textbf{Linguistic structure} is viewed as a system that generates meaningful sentences. It is commonly studied from the perspectives of phonetics (physical properties of speech sounds)~\citep{ladefoged2006course}, phonology (mental representations that govern speech patterning)~\citep{chomsky1968sound}, morphology (studies of word structure)~\citep{halle1993distributed}, and syntax (studies of sentence structure)~\citep{chomsky2002syntactic}. Through these lenses, linguists try to reveal the underlying principles of linguistic competence~\citep{pinker2003language}. On the other hand, computational phonetics captures language patterns using methods such as forced alignment~\citep{gorman2011prosodylab}, articulatory inversion~\citep{shahrebabaki2021acoustic}, and speech synthesis~\citep{triantafyllopoulos2023overview}, while phonological models simulate rule learning and constraint processing~\citep{hayes2008maximum}. Morphological models induce language structures across diverse typologies~\citep{goldsmith2001unsupervised,cotterell2016sigmorphon}. In syntax, computational approaches address parsing through tasks like sentence segmentation, text chunking and part-of-speech tagging~\citep{zhang2023syntactic,zhang2024granular}. However, current data-driven language modeling methods seem to be contrary to this explicit, modular decomposition of language. Computational methods for exploring the basic elements of language are still limited.

\textbf{Meaning} is a core concept in linguistics, including the studies in semantics and pragmatics. Semantics addresses the literal interpretation of language, assuming that meaning derives from the compositional structure of language~\citep{saeed2015semantics}. Pragmatics, in contrast, explores how meaning is influenced by context, including phenomena such as implicature~\citep{carston2008thoughts}, presupposition~\citep{potts2015presupposition}, deixis~\citep{levinson2006deixis}, and speech acts~\citep{searle2014speech}. In computational linguistics, lexical meanings are represented via either symbolic (e.g., WordNet~\citep{miller1995wordnet}, ConceptNet~\citep{speer2017conceptnet}, FrameNet~\citep{baker1998berkeley}, and PrimeNet~\citep{primenet}) or distributional approaches (e.g., Word2Vec~\citep{mikolov2013word2vec} and BERT~\citep{devlin2019bert}). Representative semantic and pragmatic processing tasks include word sense disambiguation and sarcasm detection~\citep{mao2024semantic, mao2025pragmatic}. Unlike symbolic representations that link meaningful concepts through a variety of structured relationships, distributional approaches quantify semantic similarity based on distances in vector space, offering only a surface-level approximation of meaning without capturing its underlying compositional structure.

\begin{figure}[!t]
    \centering
    \includegraphics[width=1\linewidth]{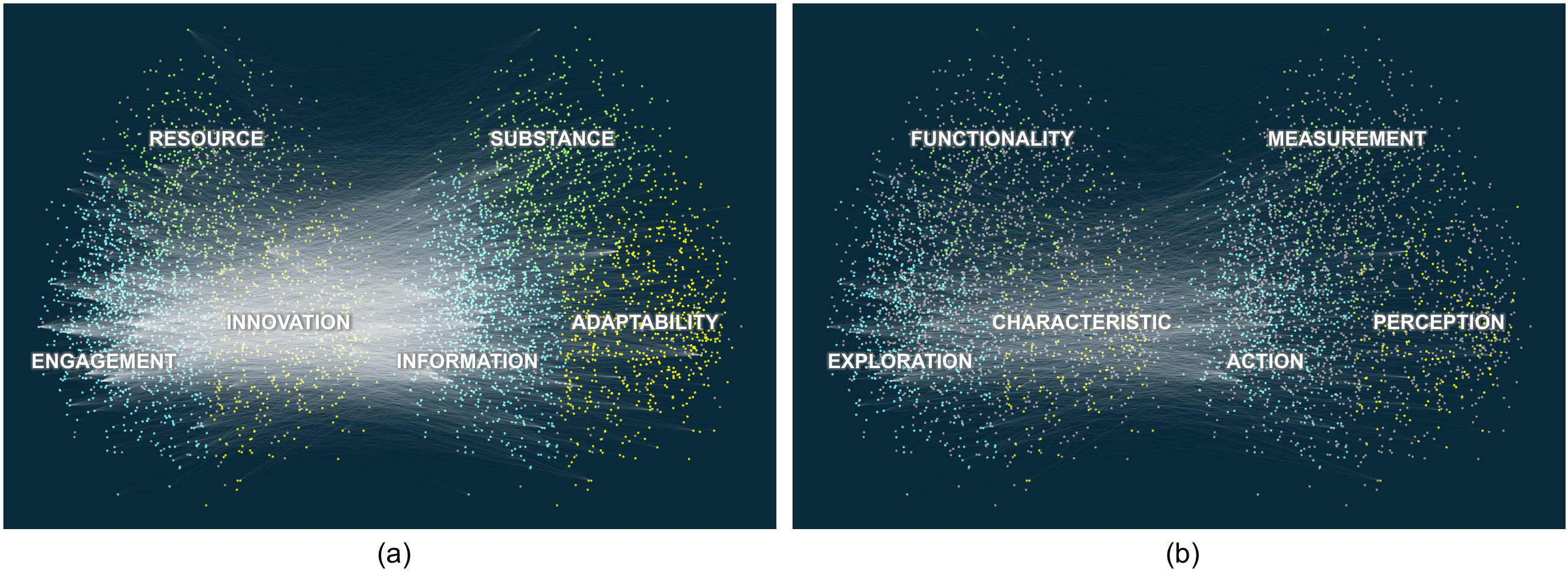}
    \caption{The metaphorical concept mappings comparison between (a) humans and (b) ChatGPT~\citep{mao2025chatgpt}. For each subject, the cluster on the left represents target concepts, while the cluster on the right represents source concepts. Gray dots indicate inactive concepts, and distinct bright colors represent different groups of activated concepts. Compared to humans' even concept distributions, ChatGPT presents distinct concept preference (more activated turquoise dots than others). Reproduced.}
    \label{fig:language}
\end{figure}

\textbf{Linguistic applications} examine language as it is used in real-world contexts. Formalist linguists argue that language acquisition depends on innate grammatical structures (Universal Grammar)~\citep{chomsky2014aspects} and is constrained by limited input (Poverty of the Stimulus)~\citep{berwick2011poverty}. In contrast, cognitive linguists believe that language learning is driven by general cognitive and socio-cognitive capacities~\citep{tomasello2005constructing}. Metaphorical language frames our cognitive systems in concept understanding~\citep{lakoff2008metaphors}. These theoretical perspectives have given rise to two distinct paradigms in natural language generation (NLG): rule-based systems (text generation with explicitly encoded grammatical and lexical rules~\citep{reiter2000building}) and language modeling-based systems (learning generation from large corpora via statistical and neural methods~\citep{bengio2003neural}). Beyond NLG, cognitive computing researchers analyzed cognitive patterns in both humans and GenAI from their metaphorical languages~\citep{mao2025chatgpt} (see Fig.~\ref{fig:language}). They also investigated how metaphors relate to neural activity~\citep{mao2025eeg}, and behavior~\citep{mao2024unveiling}. However, current AI systems lack grounding in embodied cognition, limiting their ability to capture the experiential basis of meaning.

\section{Culture}\label{sec: Culture}

Culture is the collective performance of learned behaviors, beliefs, values, attitudes, norms, traditions, rituals, and institutions that characterize a group or society, and are passed down through generations~\citep{tylor1920primitive}. Although the research in culture was categorized as seven genres, e.g., pattern (structure), function, process, product, refinement, power/ideology, or group membership~\citep{baldwin2006redefining}, cognitive scientists are mostly concerned with culture as patterns, processes, and functions~\citep{kelly2022cogsci}.

\begin{figure}[!t]
    \centering
    \includegraphics[width=0.8\linewidth]{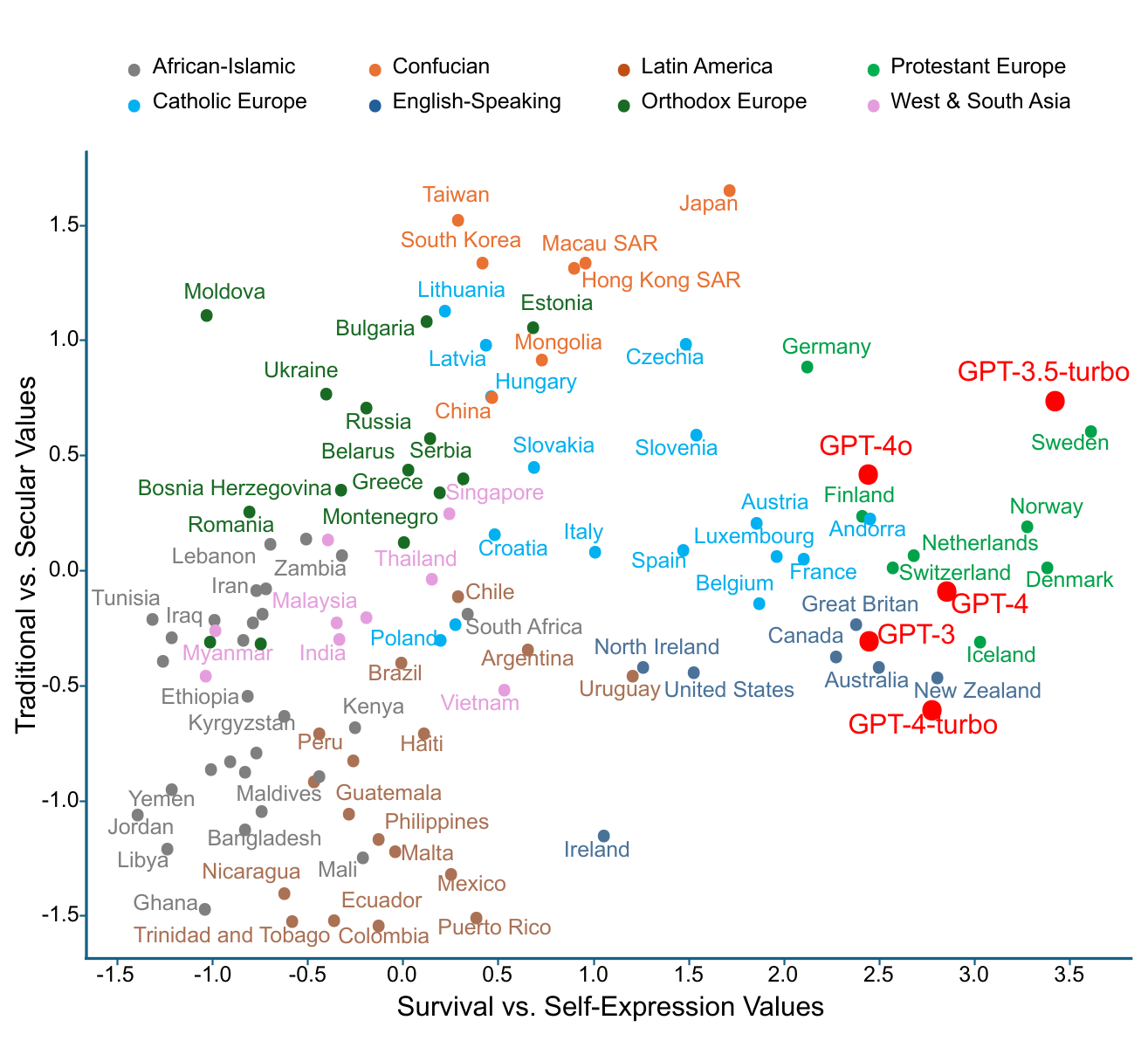}
    \caption{The map shows 107 regions from the last three waves of the Integrated Values Surveys~\citep{tao2024cultural}. The x-axis spans from survival to self-expression values, and the y-axis from traditional to secular values. Five additional red points represent responses from LLMs to the same survey questions. Reproduced.}
    \label{fig:culture}
\end{figure}

\textbf{Cultural as patterns} conceptualize culture as a persistent and structured system, such as norms, roles, and institutions, that influence social behavior. This system is stable~\citep{bourdieu2020outline} and has a hierarchical structure~\citep{gobel2024self} in social networks~\citep{granovetter1985economic}. Cross-cultural research found that East Asians tend to engage in dialectical reasoning, while Westerners emphasize rule-based logic~\citep{nisbett2001culture}. Despite cultural variation in moral expressions, there are consistent ethics that can be found in different societies, e.g., fairness, loyalty, and respect to authority~\citep{haidt2004intuitive,gray2025morality}. Recent LLM research leverages role-play to simulate individuals with different cultural backgrounds for social simulations~\citep{mou2024individual,rossetti2024social,zhu2024generative} with population-level scalability~\citep{yang2024oasis}. These simulations have been used to investigate phenomena such as partisan decision-making wisdom~\citep{chuang2024wisdom} and economics~\citep{li-etal-2024-econagent}. Although the role-play-based methods may force LLMs to generate text with certain tones, without understanding individuals' thinking frameworks, the simulations remain limited in their psychological fidelity. Furthermore, LLMs tend to bias to English-speaking, and Protestant European culture~\citep{tao2024cultural} (see Fig.~\ref{fig:culture}). These limitations highlight the need to incorporate culture as a formal and rigorous dimension in AI methodology.

\textbf{Culture as processes} views culture as a process that needs to be adapted and developed. It is continually performed~\citep{schechner2003performance}, created~\citep{shteynberg2010silent}, and transformed~\citep{blumer1986symbolic} via communication and social practices. For example, imitation, guidance, and observation can speed up the dissemination of culture, especially when a certain culture is favored~\citep{henrich2003evolution, reader2002social}. As AI systems become more socially autonomous, humans may need to adjust the view of social interactions from human-to-human to human-to-machine~\citep{matthews2021evolution}. While evolutionary algorithms (e.g., Genetic Algorithms~\citep{alhijawi2024genetic}, and Neuroevolution~\citep{stanley2002evolving}) have been developed and inspired by biogenetics, their application to modeling cultural dynamics is very limited. Instead, researchers have applied metaphor processing (MetaPro)~\citep{mao2023metaproonline} and topic modeling~\citep{blei2003latent} to examine cultural changes in different scenarios~\citep{mao2024understanding,duong2024wildfires}. Other methods for understanding culture evolution include dictionaries, e.g., Linguistic
Inquiry and Word Count~\citep{pennebaker2015liwc2015}, and PLM, because these techniques help to examine cultural expressions, and uncover parallels and evolving trends in language over time~\citep{berger2022using}. Nevertheless, the processual view of culture has not significantly influenced AI methodologies.

\textbf{Culture as functions} conceptualizes culture in terms of the roles it fulfills in society, such as maintaining social order~\citep{garland2001culture}, enhancing knowledge transmission~\citep{connelly2003predictors}, reducing uncertainty~\citep{hogg2000subjective}, and guiding individual or group behavior~\citep{smith2002cultural}. This perspective is consistent with utilitarian perspectives, which believe that culture serves critical adaptive functions related to behavioral regulation, group cohesion, and identity formation~\citep{schein2010organizational}. For instance, ``tight cultures'' enforce strict norms and exhibit a low tolerance for deviance, while ``loose cultures'' exhibit a greater degree of behavioral flexibility~\citep{gelfand2011differences}. Industrial AI exemplifies this functional view of culture. The development of smart cities positions AI as a tool for social governance, improving regulation, efficiency, and environmental forecasting~\citep{alahi2023integration}. Warehouse robots have overturned operational workflows by shifting the optimization process from human-centered to machine-driven control~\citep{qin2022jd}. Autonomous vehicles have changed both the occupational environments for drivers and public perceptions of transportation~\citep{nikitas2021autonomous}. As AI becomes more and more integrated into daily life, there is an urgent need to establish a legal framework to clearly delineate the appropriate boundaries for the deployment of AI.

\section{The Future of AI in Cognitive Science}\label{sec: The Future of AI in Cognitive Science}

\begin{table}[!tb]
\centering
\footnotesize
\caption{Maturity analysis for cognition-inspired AI (CIAI) and AI for cognitive analysis (AICA).}\label{tab: maturity}
\begin{tabular}{p{1.2cm}p{1.8cm}p{6.0cm}p{0.7cm}p{0.7cm}} 
\toprule
\multicolumn{1}{c}{\textbf{Discipline}} & \multicolumn{1}{c}{\textbf{Field}} & \multicolumn{1}{c}{\textbf{Representative AI Techniques}} & \multicolumn{1}{c}{\textbf{CIAI}} & \multicolumn{1}{c}{\textbf{AICA}} \\ 
\hline
\multirow{4}{*}{Philosophy} & Metaphysics  & GenAI, AR  VR & $\blacksquare \blacksquare \blacksquare \square$ & $\blacksquare \blacksquare \square \square$  \\
  & Epistemology & KBS, word2vec, PLM  & $\blacksquare \blacksquare \blacksquare \square$  & $\blacksquare \square \square \square$  \\
  & Logic  & GenAI, KBS, FOL, NLU, CoT, CBR  CSR & $\blacksquare \blacksquare \blacksquare \square$  & $\blacksquare \blacksquare \square \square$  \\
  & Ethics & Alignment, Bias, XAI  & $\blacksquare \blacksquare \square \square$  & $\blacksquare \square \square \square$  \\ 
\hline
\multirow{5}{*}{Psychology} & Mind & GenAI & $\blacksquare \blacksquare \blacksquare \square$  & $\blacksquare \blacksquare \square \square$  \\
  & Emotion  & Affective comput.  & $\blacksquare \blacksquare \square \square$  & $\blacksquare \square \square \square$  \\
  & Personality  & PTR, persona extract., personalized syst. & $\blacksquare \blacksquare \square \square$  & $\blacksquare \square \square \square$  \\
  & Behavior & RL, Robotics, SNA, fin. pred., behav. Interv. syst. & $\blacksquare \blacksquare \blacksquare \blacksquare$  & $\blacksquare \square \square \square$ \\
  & Mental Health  & Disorder detect. syst., XAI, GenAI  & $\blacksquare \blacksquare \blacksquare \square$  & $\blacksquare \blacksquare \square \square$  \\ 
\hline
\multirow{4}{*}{Neuroscience} & Perception & DNNs, e.g., CNNs & $\blacksquare \blacksquare \blacksquare \blacksquare$  & $\blacksquare \square \square \square$ \\
  & Memory & LSTM, GRU, knwl. bases, RAG & $\blacksquare \blacksquare \blacksquare \square$  & $\blacksquare \square \square \square$ \\
  & Attention  & Attention mechanisms  & $\blacksquare \blacksquare \blacksquare \square$  & $\blacksquare \square \square \square$ \\
  & Brain  & Multi-agents, brain signal process. & $\blacksquare \blacksquare \square \square$  & $\blacksquare \blacksquare \square \square$  \\ 
\hline
\multirow{3}{*}{Linguistics}  & Structure  & Structural ling. process.  & $\blacksquare \blacksquare \square \square$  & $\blacksquare \blacksquare \square \square$  \\
  & Meaning  & Semantic and pragmatic AI  & $\blacksquare \blacksquare \blacksquare \square$  & $\blacksquare \blacksquare \blacksquare \square$  \\
  & Applications & NLG, machine transl., cognitive comput.  & $\blacksquare \blacksquare \blacksquare \blacksquare$  & $\blacksquare \blacksquare \blacksquare \square$  \\ 
\hline
\multirow{3}{*}{Culture}  & Patterns & Role-play, social simul.  & $\blacksquare \blacksquare \square \square$  & $\blacksquare \square \square \square$ \\
  & Processes  & Gen. alg., neuroevo., MetaPro, topic model, PLM  & $\blacksquare \blacksquare \square \square$  & $\blacksquare \blacksquare \square \square$  \\
  & Functions  & Smart city, warehouse robot, auto. vehicle & $\blacksquare \blacksquare \blacksquare \square$ & $\blacksquare \square \square \square$ \\
  \hline
  \multicolumn{5}{>{}p{0.95\linewidth}}{\textbf{Maturity Levels of Cognition-Inspired AI (CIAI):}}\\
\multicolumn{5}{>{}p{0.95\linewidth}}{- Level 0. Pre-Theoretical: No AI methods are informed by principles, models, or findings from any recognized subfield of Cognitive Science. There is no conceptual or methodological bridge between cognitive theories and AI implementations.}\\
\multicolumn{5}{>{}p{0.95\linewidth}}{- Level 1. Symbolic Referencing: Cognitive Science concepts are mentioned or loosely mapped in AI models, but without substantive methodological integration. Impact is minimal and restricted to niche explorations or isolated studies within a small subset of the AI research community.}\\
\multicolumn{5}{>{}p{0.95\linewidth}}{- Level 2. Preliminary Integration: Core cognitive principles are explicitly modeled and computationally realized. These methods are recognized and evaluated within select AI subfields, though they remain peripheral to mainstream AI.}\\
\multicolumn{5}{>{}p{0.95\linewidth}}{- Level 3. Functional Deployment: Cognition-inspired models yield measurable improvements in task performance across multiple AI domains. They influence system design beyond theoretical interest, demonstrating generalizability and competitive benchmarks in practical applications.}\\
\multicolumn{5}{>{}p{0.95\linewidth}}{- Level 4. Paradigm-Level Influence: Cognition-inspired paradigms have reshaped dominant AI research agendas. These methods are foundational in both theoretical modeling and real-world AI systems, widely adopted across disciplines and often used to explain and justify AI behaviors.}\\
\multicolumn{5}{>{}p{0.95\linewidth}}{\textbf{Maturity Levels of AI for Cognitive Analysis (AICA):}}\\ 
\multicolumn{5}{>{}p{0.9\linewidth}}{- Level 0. No Relevance: AI methods are not used for cognitive analysis tasks. There is no recognized utility of AI tools in addressing Cognitive Science research questions.}\\
\multicolumn{5}{>{}p{0.95\linewidth}}{- Level 1. Limited Application: AI techniques are applied to narrowly defined cognitive tasks with limited methodological sophistication. The impact on Cognitive Science theory or practice is negligible.}\\
\multicolumn{5}{>{}p{0.95\linewidth}}{- Level 2. Exploratory Contribution: AI models are used in cognitive analysis studies. These contributions are exploratory and informative but not yet central to cognitive scientific inquiry or debates.}\\
\multicolumn{5}{>{}p{0.95\linewidth}}{- Level 3. Empirical Utility: AI methods contribute significantly to Cognitive Science by enabling new forms of empirical analysis. These tools have led to the generation of novel hypotheses or supported theory refinement in select cognitive domains.}\\
\multicolumn{5}{>{}p{0.95\linewidth}}{- Level 4. Integrated Epistemic Tool: AI techniques are deeply embedded in Cognitive Science research. They are used not only for data analysis but also as epistemic agents, e.g., tools that simulate, test, and sometimes challenge core cognitive theories. Their methodological and conceptual roles are widely recognized across the field.}  \\ 
\bottomrule
\end{tabular}
\end{table}

The future of AI in Cognitive Science encompasses not only enhancing task performance but also deepening our comprehension of the mind via systems that encapsulate the richness, adaptability, and individuality of human cognition. Despite recent advancements, particularly in GenAI, RL, and neural networks, the future necessitates a more cohesive, theory-based, and morally informed approach.

As summarized in Table~\ref{tab: maturity}, the maturity of representative AI techniques varies in terms of cognitive integration and their utility for cognitive research. While many AI models were inspired by cognitive functions (e.g., memory via LSTM, attention via Transformers, perception via CNNs), they often fail to reflect the computational purposes and representational constraints emphasized in cognitive theories. For example, although GenAI can mimic human language and tones, it lacks intentionality, embodiment, and contextual adaptability -- core attributes of human intelligence. This highlights the need to move beyond heuristic-driven engineering toward cognitive architectures, informed by areas such as neurosymbolic reasoning, developmental learning, and social cognition. Conversely, AI applications for cognitive analysis remain limited, particularly in domains like epistemology, ethics, emotion, behavioral science, and culture, where context-dependent, normative reasoning is essential. These applications often prioritize pattern recognition over theoretical modeling.

Seven major challenges and corresponding research opportunities stand out: 

\noindent 1) Aligning AI Behavior with Cognitive Frameworks: Current AI may replicate outputs without internalizing human-like reasoning. Future research should align machine outputs with interpretable cognitive models by combining symbolic and sub-symbolic methods and enhancing explainability to support theory-building.

\noindent 2) Grounding Meaning in Machines: A key challenge in AI is the symbol grounding problem -- how machines can move beyond manipulating abstract symbols to actually understanding them. Future research should ground symbols in perception, action, and interaction by linking language to sensorimotor experience, using embodied learning, and leveraging multimodal feedback. Bridging neural models with semantic networks or commonsense graphs may offer a substrate where meaning emerges through use, not just labels—advancing both AI and cognitive science.

\noindent 3) Embedding AI in Embodiment and Culture: Grounded cognition emphasizes that thought is shaped by bodily experience and cultural context. AI systems must be designed to support situated understanding, including interaction with physical environments and culturally sensitive simulation beyond Western-centric norms.

\noindent 4) Developing Individualized Cognitive Representations: Although GenAI has shown remarkable proficiency in handling varied tasks through probabilistic learning, there is a lack of unified frameworks for representing individual cognitive patterns. Future research should aim to develop cognitive representation methods that account for personalized patterns and the needs of underrepresented populations, ensuring more equitable AI systems.

\noindent 5) Integrating Multimodal and Multisensory Processing:
Human cognition seamlessly integrates inputs from vision, audition, proprioception, emotion, and more. While some models now combine text and images, they still lack the fluid, interactive processing of multisensory experience.
The next step involves building systems that understand the causal and temporal relationships between modalities, grounded in real-world interaction. These advances would support higher-level cognition such as analogical reasoning, emotional intelligence, and embodied planning—key to both AI advancement and cognitive modeling.

\noindent 6) Advancing Meta-Cognition and Self-Reflection in AI:
Humans possess the ability to monitor and regulate their own cognitive states—a capability called metacognition. This includes uncertainty estimation, strategy adaptation, and self-explanation.
Developing metacognitive AI involves designing models that can assess their own confidence, detect failure modes, and explain their reasoning to themselves and others. Such systems would be more robust, trustworthy, and cognitively transparent, enabling deeper comparisons with human introspective processes and supporting autonomous learning.

\noindent 7) Re-conceptualizing AI Ethics through Cognitive Co-evolution: As AI increasingly influences cognition via human-computer interaction, ethical frameworks must extend beyond technical compliance to encompass human flourishing. Value-sensitive design, long-term societal modeling, and interdisciplinary inquiry into meaning, autonomy, and responsibility should be foundational to AI development.

\section{Conclusion}\label{sec: Conclusion}

The convergence of AI and Cognitive Science represents a transformative frontier in both understanding the human mind and building intelligent systems. This review has mapped the interdisciplinary landscape across philosophy, psychology, neuroscience, linguistics, and culture, revealing both the methodological synergies and conceptual gaps that define current research. While AI has drawn inspiration from cognitive principles, it often abstracts away the deeper theoretical structures that ground these capacities in human experience, development, and culture. Likewise, Cognitive Science has begun to incorporate AI as a tool for simulation and analysis, yet it has not fully leveraged AI's potential for hypothesis generation, scalable experimentation, or cognitive theory examination. As our analysis of cognition-inspired AI and AI for cognitive analysis indicated, current developments are promising but uneven. Looking ahead, the future of AI for Cognitive Science depends on a shift from shallow imitation to deep modeling. It requires systems that are not only capable of performing tasks but of interpreting, adapting, and reasoning in ways that resonate with human thought and behavior. This means designing AI that is theoretically anchored, culturally attuned, ethically aware, and cognitively transparent.

\section{Acknowledgements}\label{sec: Acknowledgments}

The authors acknowledge the anonymous reviewers for the comments and suggestions. Dr.~Rui Mao and Prof.~Erik Cambria are supported by the RIE2025 Industry Alignment Fund – Industry Collaboration Projects (IAF-ICP) (Award I2301E0026), administered by A*STAR, as well as supported by Alibaba Group and NTU Singapore through Alibaba-NTU Global e-Sustainability CorpLab (ANGEL). Prof.~Hussain acknowledges the support of the UK Engineering and Physical Sciences Research Council (EPSRC) Grants Ref. EP/T021063/1 (COG-MHEAR) and EP/T024917/1 (NATGEN).

\bibliography{sn-bibliography}

\end{document}